\documentclass{article}

\usepackage[preprint]{neurips_2026}

\usepackage[utf8]{inputenc} 
\usepackage[T1]{fontenc}    
\usepackage{hyperref}       
\usepackage{url}            
\usepackage{booktabs}       
\usepackage{amsfonts}       
\usepackage{amsmath}        
\usepackage{amssymb}        
\usepackage{nicefrac}       
\usepackage{microtype}      
\usepackage{xcolor}         
\usepackage{graphicx}
\usepackage{multirow}
\usepackage{subcaption}
\usepackage{enumitem}
\usepackage{amsthm}
\usepackage{bbm}
\usepackage{float}
\makeatletter
\renewcommand{\paragraph}{\@startsection{paragraph}{4}{\z@}{0pt}{-0.5em}{\normalsize\bfseries}}
\makeatother

\newtheorem{definition}{Definition}

\newcommand{\bW}{\mathbf{W}}
\newcommand{\bX}{\mathbf{X}}
\newcommand{\bE}{\boldsymbol{\varepsilon}}
\newcommand{\bI}{\mathbf{I}}
\newcommand{\bL}{\mathbf{L}}
\newcommand{\bz}{\mathbf{z}}
\newcommand{\R}{\mathbb{R}}

\title{When Does Gene Regulatory Network Inference Break? A Controlled Diagnostic Study of Causal and Correlational Methods on Single-Cell Data}

\author{%
  Miguel Fernandez-de-Retana\thanks{Corresponding author.} \\
  Faculty of Engineering\\
  University of Deusto\\
  Bilbao, Spain \\
  \texttt{m.fernandezderetana@deusto.es} \\
  \And
  Rubén Sánchez-Corcuera\thanks{Equal contribution.} \\
  Faculty of Engineering\\
  University of Deusto\\
  Bilbao, Spain \\
  \texttt{ruben.sanchez@deusto.es} \\
  \And
  Unai Zulaika\footnotemark[2] \\
  Faculty of Engineering\\
  University of Deusto\\
  Bilbao, Spain \\
  \texttt{unai.zulaika@deusto.es} \\
  \And
  Aritz Bilbao-Jayo\footnotemark[2] \\
  Faculty of Engineering\\
  University of Deusto\\
  Bilbao, Spain \\
  \texttt{aritzbilbao@deusto.es} \\
  \And
  Aitor Almeida\footnotemark[2] \\
  Faculty of Engineering\\
  University of Deusto\\
  Bilbao, Spain \\
  \texttt{aitor.almeida@deusto.es} \\
}

\begin{document}

\maketitle

\begin{abstract}
Despite theoretical advantages, causal methods for Gene Regulatory Network (GRN) inference from single-cell RNA-seq data consistently fail to match or outperform correlation-based baselines in many realistic benchmarks, a persistent puzzle which casts doubt on the value of causality for this task. We argue that existing benchmarks are insufficiently controlled to answer this question because they evaluate on real or semi-real data where multiple pathologies co-occur, confounding failure modes, and obscuring the specific conditions under which different inference methods excel or fail. To address this gap, we introduce a controlled diagnostic framework that isolates seven biologically motivated pathologies (dropout, latent confounders, cell-type mixing, feedback loops, network density, sample size, and pseudotime drift) and measure how six representative methods spanning three inference paradigms degrade as each pathology intensifies. Across 6,120 controlled experiments, we find that causal methods genuinely dominate in clean and structurally favorable regimes, but specific pathologies (notably dropout and latent confounders) selectively neutralize their advantages. We further introduce an error-type decomposition that reveals methods with similar aggregate accuracy commit qualitatively different errors. To probe whether single-pathology effects persist when multiple stressors co-occur, we perform an interaction sweep over the three most impactful pathologies and find that their joint effects are sub-additive, while also exposing density-conditional cross-overs invisible to single-dial analysis. Our findings offer a nuanced understanding of when and why different methods succeed or fail for GRN inference, providing actionable insights for method development and practical guidance for practitioners.\footnote{Code available at: \url{https://github.com/miguelfrndz/GRN-Inference-Diagnosis}}
\end{abstract}


\section{Introduction}
\label{sec:intro}

Inferring Gene Regulatory Networks (GRNs) from single-cell transcriptomic data is a central problem in computational biology, with profound implications for understanding cellular function, disease mechanisms, drug response, and therapeutic interventions, among others \citep{kamimoto2023dissecting,aibar2017scenic}. Fundamentally, a GRN is a \textit{directed} graph $\mathcal{G} = (V, E)$ encoding \textit{regulatory relationships} where nodes ($V$) represent genes and edges ($E$) represent functional interactions, such as transcription factors activating or repressing target genes in specific cellular contexts. More formally, the problem of GRN inference can be framed as a structure learning task, where given expression data for $p$ genes across $n$ cells, the goal is to recover the directed adjacency matrix $\bW \in \R^{p \times p}$ of the underlying regulatory graph $\mathcal{G}$, where $W_{ij} \neq 0$ indicates a regulatory influence of gene $i$ on gene $j$ and the sign of $W_{ij}$ indicates activation or repression.

Historically, a wide variety of methods have been proposed for GRN inference, ranging from simple correlation-based approaches (e.g., Pearson correlation, mutual information) to more sophisticated causal inference methods (e.g., PC, GES, NOTEARS) that explicitly model the data-generating process and attempt to recover directed edges. Theoretically, causal methods should have a fundamental advantage: by modeling the underlying generative process, they can, in principle, distinguish direct regulatory interactions from indirect associations and confounding effects, leading to more accurate and interpretable GRN reconstructions. However, a growing body of empirical evidence reveals a puzzling finding: simple correlation-based methods such as Pearson correlation and GENIE3 \citep{huynh2010inferring} frequently match or outperform theoretically stronger causal inference methods on real single-cell data. This has been documented systematically by CausalBench \citep{chevalley2025causalbench}, which found that causal methods do not reliably outperform baselines on perturbational benchmarks, and by geneRNIB \citep{nourisa2025genernib}, whose living benchmark reported that simple correlational models often outperformed more complex approaches. These findings have led some to question whether the additional complexity of causal methods is justified for GRN inference at all.

We argue that this conclusion is premature. Existing benchmarks evaluate on real or semi-real data where multiple data pathologies, understood as biological or technical challenges that violate the assumptions of different inference methods, co-occur: dropout artifacts, latent confounders, cell-type heterogeneity, and non-stationarity are all present simultaneously. Because these pathologies are confounded with each other and with the unknown ground-truth, benchmarks are \emph{underpowered} to attribute failures to specific causes. In other words, observing that method $\mathcal{A}$ beats method $\mathcal{B}$ on dataset $\mathcal{D}$ does not tell us \emph{why}, because we do not know which pathology (or combination of pathologies) is responsible for the performance gap. This lack of diagnostic resolution means we cannot identify the specific conditions under which different methods excel or break, and without knowing why, we cannot improve methods or advise practitioners.

\paragraph{Our Approach.}
We introduce a controlled diagnostic framework inspired by ablation studies in Machine Learning (ML) and perturbation-based analyses to isolate method failure mechanisms under specific pathologies and quantify their impact on performance. We construct a synthetic \textit{simulator} building upon the previous work by \citet{dibaeinia2020sergio} that generates single-cell expression data from a known linear structural causal model with seven independently controllable pathology \textit{``dials''}. Each dial corresponds to a specific biological or technical challenge: dropout, latent confounders, cell-type mixing, feedback loops, network density, sample size, and non-stationarity along a pseudotime trajectory (pseudotime drift). By sweeping one dial at a time while holding all others fixed, we isolate the mechanism of failure for each pathology and quantify its impact on performance. This approach allows us to answer questions such as: \emph{Does dropout selectively neutralize the advantage of causal methods? Do latent confounders degrade all methods equally? Are some methods more robust to feedback loops than others?} By systematically characterizing the degradation patterns of six representative methods spanning three inference paradigms (correlational, tree-ensemble, and causal), we provide a thorough understanding of when and why methods succeed or fail for GRN inference. 

We evaluate six representative methods spanning five method classes: correlational (Pearson, mutual information), tree-ensemble (GENIE3), constraint-based causal (PC), score-based causal (GES), and optimization-based causal (NOTEARS); using two complementary metrics: Area Under the Precision--Recall Curve (AUPRC) for characterizing the edge-recovery accuracy of each method as pathology intensifies, and a novel \emph{error-type decomposition} that classifies each predicted edge as true, reversed, confounded, spurious, or missed.

\paragraph{Key Findings.} Across 2,100 controlled experiments ($7$ pathologies $\times$ $5$ levels $\times$ $10$ replicates $\times$ $6$ methods) for both linear and nonlinear SCMs, and a complementary interaction sweep over dropout, latent confounders, and density, we find that no method breaks uniformly: causal methods (NOTEARS, GES) dominate under clean conditions (AUPRC $> 0.94$) and remain the top performer at the hardest level of 6 of the 7 pathology dimensions, but heavy dropout is the exception, where Pearson correlation outperforms all methods. Dropout is also the most discriminating: MI and GENIE3 collapse ($\Delta$AUPRC $\approx -0.7$), while Pearson degrades gracefully ($\Delta$AUPRC $= -0.28$). Latent confounders and cell-type mixing are the great equalizers, compressing all methods into narrow bands at comparable rates. Error decomposition shows that similar aggregate AUPRC can hide qualitatively different behaviors (e.g., methods may differ in whether their errors are orientation reversals, confounder-induced false positives, or pure hallucinations). Finally, the interaction sweep shows that joint pathologies degrade performance markedly less than the sum of their individual effects, while also revealing density-conditional cross-overs invisible to single-dial analysis.

\section{Related Work}
\label{sec:related}

\paragraph{GRN Inference Benchmarks.} BEELINE \citep{pratapa2020benchmarking} benchmarked 12 algorithms on synthetic (BoolODE) and curated networks, establishing AUROC and AUPRC as standard metrics for GRN evaluation. The DREAM3-5 challenges \citep{prill2010towards,marbach2012wisdom} systematically evaluated methods on simulated multi-factorial perturbation data. More recently, CausalBench \citep{chevalley2025causalbench} introduced a large-scale benchmark using real CRISPR single-cell perturbation data with biologically grounded metrics, and geneRNIB \citep{nourisa2025genernib} proposed a living benchmark with eight causal-inference metrics across five diverse datasets, with a public leaderboard for continuous evaluation. In both cases, correlation-based methods were found to perform on par with, or better than, more advanced causal methods. Our work is complementary: rather than adding another benchmark on real data, we go in the opposite direction and ask \emph{why} methods fail, using controlled simulations where the data-generating process is known and specific pathologies can be isolated.

\paragraph{GRN Inference Methods.} GENIE3 \citep{huynh2010inferring} and its accelerated variant GRNBoost2 \citep{moerman2019grnboost2} remain dominant baselines, forming the backbone of the SCENIC pipeline \citep{aibar2017scenic, bravo2023scenicPlus}. Recent Deep Learning (DL) approaches include DeepSEM \citep{shu2021modeling}, which parameterizes the GRN as a structural equation model within a Variational Autoencoder (VAE), and single-cell foundation models such as scGPT \citep{cui2024scgpt}, GeneFormer \citep{theodoris2023transfer}, and scPRINT \citep{kalfon2025scprint}. While these methods have shown promising results, a critical evaluation by \citet{ahlmann2025deep} found that DL-based perturbation prediction does not yet outperform simple linear baselines on well-calibrated metrics.

\paragraph{Causal Structure Learning.}
Constraint-based methods such as Peter--Clark (PC) \citep{spirtes2000causation} test conditional independences to reconstruct a graph skeleton, while score-based methods such as Greedy Equivalence Search (GES) \citep{chickering2002optimal} search over equivalence classes by optimizing a consistent scoring criterion (e.g., BIC). NOTEARS \citep{zheng2018dags} reformulated DAG learning as continuous optimization with an algebraic acyclicity constraint, spawning a family of differentiable methods including DAGMA \citep{bello2022dagma}, DAG-GNN \citep{yu2019dag}, GraN-DAG \citep{lachapelle2020gradient}, DCDI \citep{brouillard2020differentiable}, DCD-FG \citep{lopez2022large}, and SDCD \citep{nazaret2024stable}. In this work, we focus on the original NOTEARS formulation as a representative optimization-based method, and GES as a representative score-based method, to provide a clear contrast with correlational baselines; our diagnostic contribution is orthogonal to architectural improvements.

\paragraph{Perturbation-Based Evaluation.}
Genome-scale Perturb-seq \citep{replogle2022mapping} has made interventional single-cell data widely available at the scale needed to evaluate directed regulatory hypotheses, and has thus become a natural ground-truth for GRN inference evaluation. This has enabled two related but distinct evaluation paradigms: network-inference benchmarks such as CausalBench, which compare inferred edges against perturbational evidence, and perturbation-prediction methods such as GEARS \citep{roohani2024predicting} and CellOracle \citep{kamimoto2023dissecting}, which evaluate whether a model can predict transcriptional responses under held-out interventions. However, predictive success under perturbation is not equivalent to causal structure recovery: a model may predict average expression shifts without identifying direct regulators, and recent critiques show that complex perturbation predictors can fail to outperform simple linear baselines under well-calibrated metrics \citep{ahlmann2025deep}. Conversely, perturbational benchmarks provide realistic validation but still entangle biological heterogeneity, measurement noise, intervention strength, and target-selection biases, which is precisely the entanglement our controlled framework seeks to disentangle.

\section{Diagnostic Framework}
\label{sec:framework}

\subsection{Data Generating Process}
\label{sec:dgp}

Our goal is not to simulate every feature of single-cell biology, but to build a diagnostic test bed in which the ground-truth regulatory graph is known and individual failure mechanisms can be switched on independently. We use a linear additive-noise Structural Causal Model (SCM) \citep{pearl2009causality} as the primary data-generating process and a common baseline for GRN inference. This choice is deliberately favorable to causal discovery methods such as NOTEARS and GES, whose assumptions are closest to this model; consequently, any degradation they exhibit cannot be attributed to an unfairly mismatched simulator. Lastly, to check that our conclusions are not artifacts of linearity, we repeat the full pathology sweep under a nonlinear SCM in Appendix \ref{app:nonlinear}, where the qualitative degradation patterns are preserved.

\paragraph{Ground-Truth Graph.}
Given $p$ genes, we sample a directed acyclic graph (DAG) $\mathcal{G} = (V, E)$ using the natural node order $0 \to 1 \to \cdots \to p-1$ as the topological order. For each target gene $j$, the number of parents is sampled as $m_j \sim \text{Binomial}(j, \rho)$, where $\rho$ controls the expected density. Conditional on $m_j$, parents are sampled without replacement from $\{0,\ldots,j-1\}$ using normalized selection weights $q_{ij}$ as in \eqref{eq:parent_selection}:
\begin{equation}\label{eq:parent_selection}
    q_{ij} = \frac{(i+1)^{-1}}{\sum_{\ell=0}^{j-1}(\ell+1)^{-1}}, \qquad i < j
\end{equation}
so, in practice, earlier genes are more likely to act as regulators. This is a \textit{parent-count} model with \textit{biased} parent selection. Edge signs are sampled uniformly from $\{-1, +1\}$ and magnitudes from $\mathcal{U}(0.5, 1.0)$, giving $W_{ij} \in [-1,-0.5] \cup [0.5,1]$ for selected edges.

\paragraph{Linear SCM.}
Conditioned on the graph, each cell $c \in \{1, \ldots, n\}$ is an independent draw from the SCM, where $X_j^{(c)} \in \R$ is the simulated expression level of gene $j$ in cell $c$:
\begin{equation}
\label{eq:scm}
    X_j^{(c)} = \sum_{i \in \mathcal{P}_\mathcal{G}(j)} W_{ij}\, X_i^{(c)} + \varepsilon_j^{(c)}, \quad \varepsilon_j^{(c)} \overset{\text{iid}}{\sim} \mathcal{N}(0, \sigma^2)
\end{equation}
where $\mathcal{P}_\mathcal{G}(j)$ denotes the functional parents of gene $j$ in $\mathcal{G}$ and $\varepsilon_j^{(c)}$ is the exogenous noise term, modeling unobserved factors and intrinsic stochasticity. The noise variance $\sigma^2$ controls the signal-to-noise ratio (SNR) of the data, with higher $\sigma$ making inference more challenging.

In matrix form, taking $\bX \in \R^{n \times p}$ as the data matrix of $n$ cells and $p$ genes, $\bW \in \R^{p \times p}$ as the weighted adjacency matrix, and $\bE \in \R^{n \times p}$ as the noise matrix, the linear SCM can be expressed as:
\begin{equation}
\label{eq:scm_matrix}
    \bX = \bE\, (\bI - \bW)^{-1}
\end{equation}
where $\bI$ is the identity matrix. In the default DAG setting this transformation encodes the cumulative effects of all directed paths; when the feedback pathology is enabled, back-edges are added after the DAG is sampled and rescaled if needed for numerical stability, but evaluation still uses the original acyclic graph as the directed ground-truth.

\subsection{Pathology Dimensions}
\label{sec:pathologies}

Each experiment activates one \textit{pathology} dial, while keeping the remaining dials at their (benign) defaults. Unless a dial explicitly changes it, the simulator uses $p = 25$ genes, $n = 800$ cells, density $\rho = 0.1$, noise scale $\sigma = 1.0$, no dropout, no latent confounders, no cell-type mixing, no feedback, and no pseudotime drift. We define the seven pathology dimensions as follows:

\paragraph{P1. Dropout ($\delta \in [0, 0.8]$).}
After expression is generated, each entry is zeroed with an \emph{expression-dependent} probability, as in \eqref{eq:dropout}:
\begin{equation}\label{eq:dropout}
    \tilde{X}_j^{(c)} = X_j^{(c)} B_j^{(c)}, \quad B_j^{(c)} \sim \text{Bernoulli}\!\left(1 - \exp\left[-\lambda\,(X_j^{(c)} - X_{\min})\right]\right)
\end{equation}
where $X_{\min}$ is the global expression minimum across all cells and genes, and $\lambda > 0$ is calibrated via binary search so that the marginal dropout rate $\mathbb{E}[1 - B_j^{(c)}] = \delta$ for $\delta \in \{0, 0.2, 0.4, 0.6, 0.8\}$. This models the dropout phenomenon in single-cell RNA-seq, where lowly expressed genes are often not detected, leading to zero-inflated data. 

\paragraph{P2. Latent Confounders ($k \in \{0, 2, 4, 8, 16\}$).}
We add $k$ unobserved Gaussian factors to the exogenous noise terms as in \eqref{eq:confounders}:
\begin{equation}\label{eq:confounders}
    \varepsilon_j^{(c)} \leftarrow \varepsilon_j^{(c)} + \bL_j^\top \bz^{(c)}, \quad \bz^{(c)} \sim \mathcal{N}(\mathbf{0}, \bI_k)
\end{equation}
where $\bL \in \R^{p \times k}$ is a sparse \textit{loading} matrix and has entries sampled from $\mathcal{N}(0,1)$ and then masked independently so that each confounder loads on approximately 30\% of genes, modeling the effect of unobserved hidden confounding factors.

\paragraph{P3. Cell-Type Mixing ($\alpha \in \{0, 0.1, 0.25, 0.4, 0.5\}$).}
We draw a fraction $(1-\alpha)$ of cells from the primary SCM and a fraction $\alpha$ from a second SCM with an independently sampled graph as in \eqref{eq:mixing}:
\begin{equation}\label{eq:mixing}
    \bX = \begin{pmatrix} \bX_A \\ \bX_B \end{pmatrix}, \quad |\bX_A| = (1-\alpha)n, \quad |\bX_B| = \alpha n
\end{equation}
where $\bX_A \sim \text{SCM}(\bW_1)$ and $\bX_B \sim \text{SCM}(\bW_2)$. The mixture is shuffled before inference, and evaluation is performed, as in all cases, against the primary graph $\bW_1$.

\paragraph{P4. Feedback Loops ($\phi \in \{0, 0.1, 0.2, 0.3, 0.5\}$).}
For each edge in the base DAG, a reverse edge is added with probability $\phi$.
Back-edge signs are sampled uniformly from $\{-1,+1\}$ and magnitudes from $\mathcal{U}(0.1,0.3)$, making feedback weaker than the original DAG edges. If the resulting weighted graph has spectral radius at least 0.9, all weights are rescaled by $0.85 / \rho_{\mathrm{spec}}(\bW)$ for numerical stability.

\paragraph{P5. Network Density ($\rho \in \{0.05, 0.1, 0.15, 0.2, 0.3\}$).}
We vary the density parameter used in the parent-count model, $m_j \sim \text{Binomial}(j,\rho)$. Thus, $\rho$ controls the expected sparsity of the base DAG and the average number of parents per gene, which in turn affects the complexity of the inference task. Sparser graphs are generally easier to infer, while denser graphs with more parents per gene create more complex dependencies.

\paragraph{P6. Sample Size ($n \in \{200, 400, 800, 1600, 3200\}$).}
We vary the number of observed cells $n$, thus isolating the statistical-power effect from changes in graph complexity.

\paragraph{P7. Non-Stationary Pseudotime Drift ($\tau \in \{0, 0.2, 0.5, 1.0, 1.5\}$).}
We partition cells into 10 equal chunks along pseudotime $t \in [0, 1]$ and generate each chunk from a time-varying SCM $\bW(t)$ as in \eqref{eq:pseudotime} where $\tau$ controls the degree of non-stationarity.
\begin{equation}\label{eq:pseudotime}
    \bW(t) = \bW \cdot \bigl(1 + \tau\,(t - 0.5)\bigr)
\end{equation}

\subsection{Inference Methods}
\label{sec:methods}

We evaluate six representative methods spanning correlational, tree-based, constraint-based, score-based, and continuous-optimization paradigms. Each returns a score matrix $\mathbf{S} \in \mathbb{R}^{p \times p}$, where larger $S_{ij}$ indicates stronger evidence for $i \to j$. We use default hyperparameters and standard Python implementations to preserve the diagnostic focus on degradation patterns under controlled pathologies, rather than optimizing performance through hyperparameter tuning, which would confound our ability to attribute failures to specific causes.


\textbf{Pearson correlation} computes $S_{ij}=|\mathrm{corr}(X_{\cdot i},X_{\cdot j})|$, yielding symmetric association scores. \textbf{Mutual information (MI)} estimates pairwise MI after equal-frequency discretization into 6 bins \citep{cover1991elements}, also producing symmetric scores. \textbf{GENIE3} \citep{huynh2010inferring} regresses each target gene on all others using a Random Forest with 50 trees and feature importances as asymmetric edge scores. \textbf{Peter--Clark (PC)} \citep{spirtes2000causation} performs Fisher $z$ conditional-independence tests up to conditioning-set size 2, returning the skeleton weighted by marginal correlation. \textbf{Greedy Equivalence Search (GES)} \citep{chickering2002optimal} uses a greedy forward BIC search with at most 3 parents per node and a variance-based proxy topological order. \textbf{NOTEARS} \citep{zheng2018dags} solves the penalized linear least-squares objective in \eqref{eq:notears} with algebraic acyclicity constraint $h(\bW)$, then thresholds small coefficients and returns absolute edge weights.
\begin{equation}\label{eq:notears}
    \begin{aligned}
    \min_{\bW} &\quad \frac{1}{2n} \|\bX - \bX\bW\|_F^2 + \lambda \|\bW\|_1 \\
    \text{s.t.} &\quad h(\bW) = \text{tr}(e^{\bW \circ \bW}) - p = 0
    \end{aligned}
\end{equation}

\subsection{Evaluation Metrics}
\label{sec:metrics}

All methods are evaluated against the original DAG $\mathcal{G}$ and under both the linear and nonlinear SCMs. We use two complementary metrics to characterize performance degradation as pathology intensifies.

\paragraph{AUPRC.}
For the \textbf{undirected AUPRC}, we collapse edge direction by taking $S_{ij}^{\mathrm{sym}} = \max(S_{ij}, S_{ji})$ and $A_{ij}^{\mathrm{sym}} = \mathbbm{1}\{A_{ij} + A_{ji} > 0\}$. Precision--recall is then computed over the upper-triangular entries. This is the primary metric for comparing all methods because it does not penalize symmetric methods for lacking orientation. For the \textbf{directed AUPRC}, we compute precision--recall over all off-diagonal ordered pairs, treating $(i \to j)$ and $(j \to i)$ as distinct predictions. This metric is computed for every method; symmetric methods are not excluded, but their tied scores in both directions naturally limit directional recovery.

\paragraph{Error-Type Decomposition.}
Given a threshold at the top-$K$ scored edges, where $K = \sum_{ij} A_{ij}$ is the number of ground-truth directed edges, we classify each predicted edge $(i, j)$ into one of five categories relative to the ground-truth DAG $\mathcal{G}$:
\begin{align}
    \textbf{True}:& \quad (i, j) \in E(\mathcal{G}) \label{eq:err_true}\\
    \textbf{Reversed}:& \quad (j, i) \in E(\mathcal{G}) \text{ but } (i,j) \notin E(\mathcal{G}) \label{eq:err_rev}\\
    \textbf{Confounded}:& \quad i,j \text{ share a common ancestor in } \mathcal{G} \text{ but no direct edge} \label{eq:err_conf}\\
    \textbf{Missed}:& \quad (i, j) \in E(\mathcal{G}) \text{ but } (i,j) \text{ not in top-$K$} \label{eq:err_miss}\\
    \textbf{Spurious}:& \quad \text{none of the above} \label{eq:err_spur}
\end{align}
This decomposition reveals whether a method \textit{fails} by reversing directions, hallucinating confounder-induced associations, or producing entirely spurious predictions, distinctions that aggregate AUPRC can mask. Two methods with the same AUPRC may therefore require different methodological fixes.

\section{Experiments and Results}
\label{sec:results}

We run $7$ pathology sweeps $\times$ $5$ severity levels $\times$ $10$ replicates $\times$ $6$ methods, for $2{,}100$ linear SCM experiments, repeat the grid under a nonlinear SCM in Appendix \ref{app:nonlinear}, and complement these single-dial sweeps with an interaction analysis over the three most informative pathologies in Section \ref{sec:interactions}. The main text emphasizes undirected AUPRC because it compares all methods on edge recovery without penalizing symmetric scores for missing orientation; directed AUPRC and runtime results are reported in the Appendix.

\subsection{Degradation Curves}
\label{sec:degradation}

\begin{figure}[t]
    \centering
    \includegraphics[width=\textwidth]{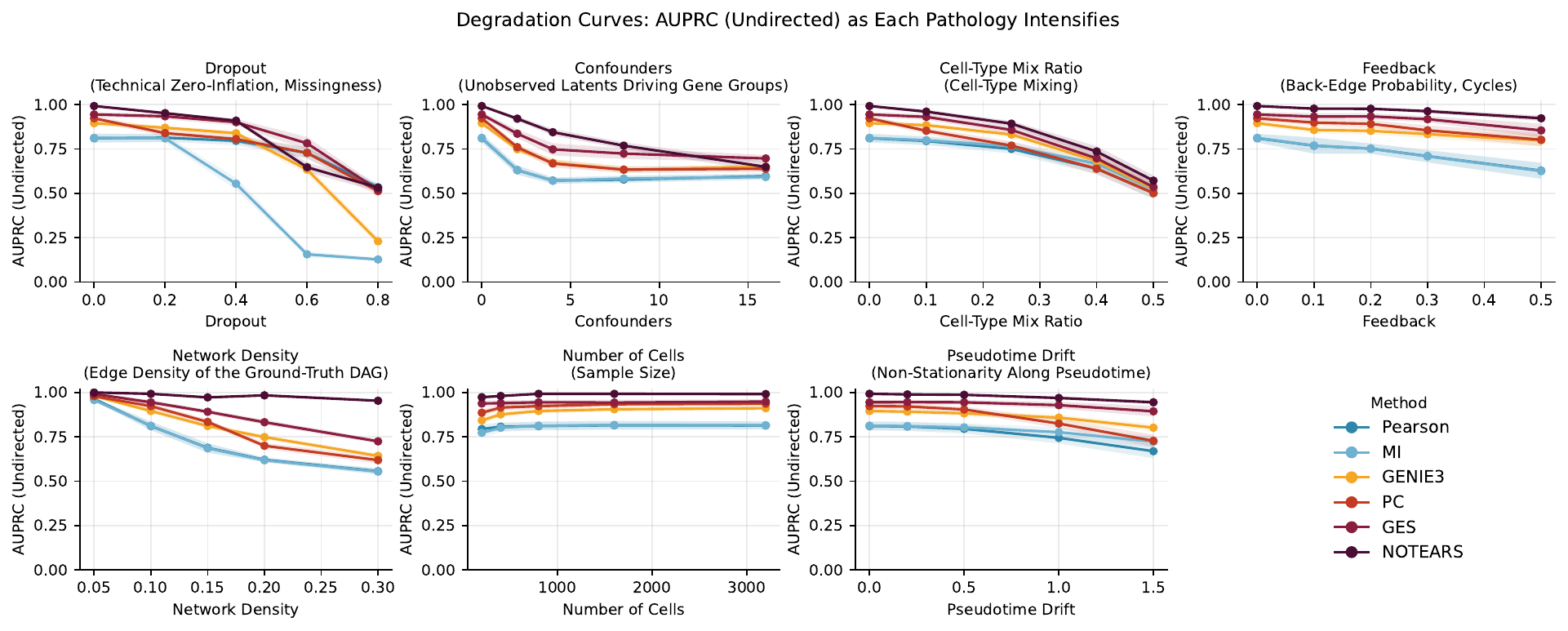}
    \caption{Undirected AUPRC as each pathology intensifies. Lines show mean $\pm$ SEM over 10 seeds. Causal methods (GES, NOTEARS) dominate under clean conditions (leftmost points), but degradation is pathology-specific: MI and GENIE3 are most fragile under dropout, confounding and cell-type mixing narrow method differences, and NOTEARS is most robust to feedback, density, sample size, and pseudotime drift.}
    \label{fig:headline}
\end{figure}

Figure \ref{fig:headline} shows that the clean setting is not the hard case: under the default linear SCM, structural methods dominate, with NOTEARS at $0.992$ undirected AUPRC and GES at $0.944$, compared with $0.811$--$0.895$ for Pearson, MI, and GENIE3. The diagnostic value comes from how this ordering changes as each pathology intensifies, revealing strikingly different degradation profiles across methods and pathologies.

\paragraph{Baseline Performance.}
Under clean conditions (leftmost points in each panel), NOTEARS achieves near-perfect recovery ($0.992$ AUPRC), followed by GES ($0.944$), PC ($0.923$), GENIE3 ($0.895$), MI ($0.813$), and Pearson ($0.811$). Explicitly structural methods with directed output (GES, NOTEARS) dominate, while symmetric correlational methods remain competitive above $0.8$ AUPRC.
\paragraph{Where Each Method Breaks.}
Table \ref{tab:delta} summarizes the change from the easiest to hardest level of each sweep, and three patterns make the failure modes mechanistically explicit. First, \emph{dropout breaks discretization- and tree-based scores most aggressively}: MI and GENIE3 collapse by $0.69$ and $0.67$ AUPRC, while Pearson, PC, GES, and NOTEARS converge into a narrow $0.51$--$0.53$ band at $\delta=0.8$ (Pearson $0.53$, NOTEARS $0.53$, GES $0.52$, PC $0.51$). At this level, Pearson is the nominal best by a margin smaller than its standard error, so the operative finding is that heavy dropout neutralizes structural advantages rather than reversing them. Second, \emph{latent confounding and cell-type mixing break every method class at comparable rates}: at the hardest levels, all six methods fall into narrow bands ($0.59$--$0.70$ under confounding; $0.50$--$0.57$ under mixing), so no method class is more robust than the others to these pathologies. Third, \emph{NOTEARS is the only method that does not appreciably break under structural pathologies}: it remains at $0.923$ under feedback, $0.953$ under high density, $0.991$ at the largest sample size, and $0.944$ under pseudotime drift, all settings where correlational approaches lose $0.09$--$0.41$ AUPRC.

\begin{table}[t]
\centering
\caption{AUPRC (undirected) drop from baseline to hardest pathology level. \textbf{Bold} indicates the most robust method per pathology (smallest $|\Delta|$); \underline{underline} indicates the most fragile (largest $|\Delta|$). For sample size, positive $\Delta$ (improvement) is expected; bold marks the largest gain.}
\label{tab:delta}
\vspace{0.5em}
\small
\setlength{\tabcolsep}{4.5pt}
\begin{tabular}{@{}lccccccc@{}}
\toprule
\textbf{Pathology} & \textbf{Range} & \textbf{Pearson} & \textbf{MI} & \textbf{GENIE3} & \textbf{PC} & \textbf{GES} & \textbf{NOTEARS} \\
\midrule
Dropout        & $0 \to 0.8$  & $\mathbf{-0.28}$ & $\underline{-0.69}$ & $-0.67$ & $-0.41$ & $-0.43$ & $-0.46$ \\
Confounders    & $0 \to 16$   & $\mathbf{-0.21}$ & $-0.22$ & $-0.25$ & $-0.28$ & $-0.25$ & $\underline{-0.34}$ \\
Cell-type mix  & $0 \to 0.5$  & $-0.31$ & $\mathbf{-0.29}$ & $-0.37$ & $\underline{-0.42}$ & $-0.41$ & $-0.42$ \\
Feedback       & $0 \to 0.5$  & $-0.18$ & $\underline{-0.19}$ & $-0.10$ & $-0.12$ & $-0.09$ & $\mathbf{-0.07}$ \\
Density        & $0.05\!\to\!0.3$ & $\underline{-0.41}$ & $-0.40$ & $-0.34$ & $-0.36$ & $-0.27$ & $\mathbf{-0.05}$ \\
Sample size    & $200\!\to\!3200$ & $+0.02$ & $+0.04$ & $\mathbf{+0.07}$ & $+0.05$ & $+0.01$ & $+0.02$ \\
Pseudotime     & $0 \to 1.5$  & $-0.14$ & $-0.09$ & $-0.09$ & $\underline{-0.20}$ & $-0.05$ & $\mathbf{-0.05}$ \\
\bottomrule
\end{tabular}
\end{table}

\subsection{Error-Type Analysis}
\label{sec:error_analysis}

\begin{figure}[t]
    \centering
    \includegraphics[width=\textwidth]{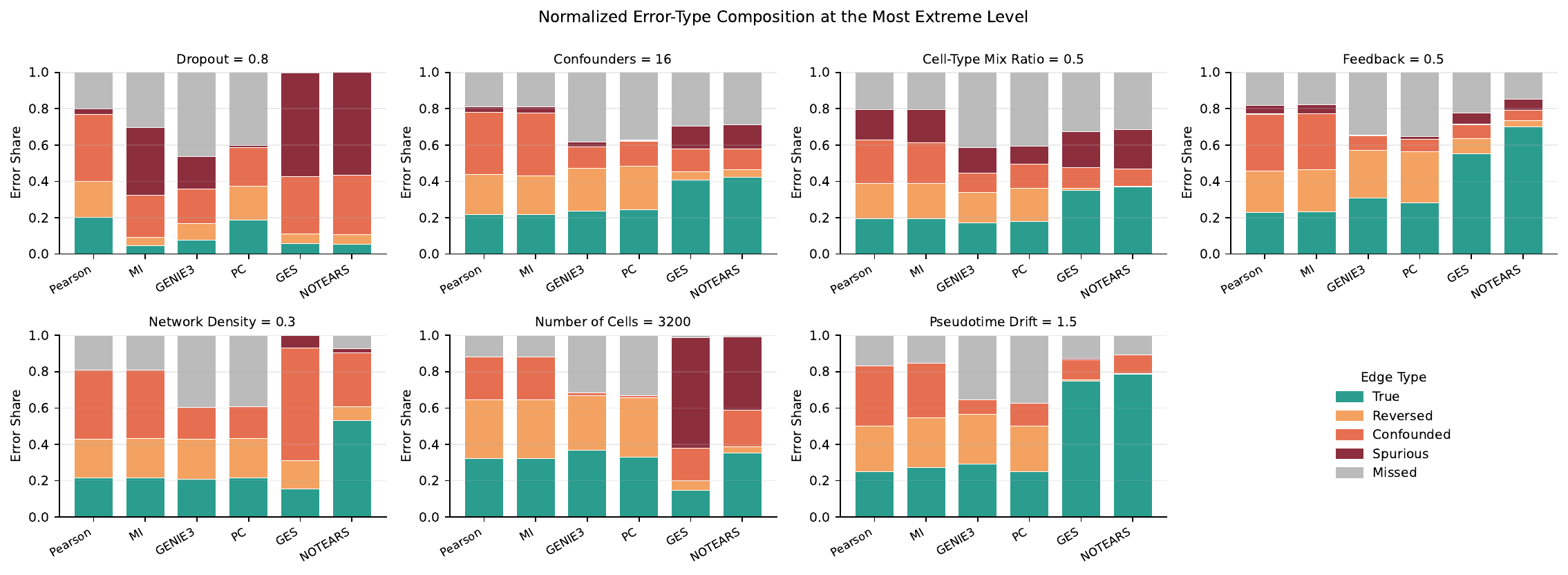}
    \caption{Normalized error-type decomposition at the hardest level of each pathology. Each bar shows the proportion of selected or missed edges assigned to each category. Structural methods recover more true directed edges in several regimes, but their failures are often confident false positives: NOTEARS and GES have more confounded and spurious errors, while Pearson and MI have more missed and reversed edges. NOTEARS maintains the highest \textbf{True} count across most pathologies.}
    \label{fig:error_decomp}
\end{figure}

Aggregate AUPRC hides qualitatively different failure modes. Figure \ref{fig:error_decomp} shows the normalized error-type decomposition at the hardest level of each pathology into true, reversed, confounded, spurious, and missed edges. Three structural patterns stand out.

\paragraph{Directional Reliability.} The reversal rate, i.e., the fraction of top-$K$ predictions that are backward-oriented, is the sharpest separator between structural and correlational methods. NOTEARS and GES maintain reversal rates below 16\% across all seven pathologies, while symmetric methods (Pearson, MI, PC) consistently show reversal rates that match their true-positive rates. The contrast is most striking under feedback ($\phi = 0.5$): NOTEARS achieves about $70\%$ true and $3\%$ reversed, while Pearson scores about $23\%$ true and $23\%$ reversed, a result of its inability to distinguish causal direction.

\paragraph{Hallucination Versus Collapse Under Dropout.} At $\delta = 0.8$, NOTEARS and GES fill all $K$ prediction slots but with 57\% spurious edges, while MI and GENIE3 produce an abundance of spurious edges and near-empty effective prediction sets. 

\paragraph{Confounded Inflation Under Density.} In the dense regime ($\rho = 0.3$), GES accumulates about $62\%$ confounded predictions at near-zero missed edges, while NOTEARS reaches $53\%$ true, $30\%$ confounded, and $8\%$ reversals. This divergence reflects differing responses to co-expression: GES includes all high-scoring candidate edges and becomes flooded by indirect associations, whereas NOTEARS's $\ell_1$ sparsity penalty preferentially selects a smaller subset of edges that are more likely to be direct, thus maintaining a higher true-positive rate at the cost of some missed edges.

\subsection{Pathology-Specific Diagnoses}
\label{sec:diagnoses}

We now synthesize the degradation curves (Figure \ref{fig:headline}) and error decomposition (Figure \ref{fig:error_decomp}) into mechanistic explanations for each pathology.

\paragraph{Why MI and GENIE3 Collapse Under Dropout.}
MI relies on equal-frequency discretization into 6 bins, and at $\delta=0.8$ most observed entries are forced to zero. This severely distorts the entropy estimate, and thus MI collapses to near-random performance. GENIE3's random forests become similarly uninformative when $\sim$80\% of feature values are zero, as splits cannot reliably separate signals from noise. Pearson correlation, by contrast, is computed over all observations (including zeros) and empirically preserves more of the pairwise edge-score ranking under MNAR zero-inflation, even as magnitudes are attenuated.

\paragraph{Why Latent Confounders Degrade All Methods Uniformly.}
The causal sufficiency assumption, that all common causes of measured variables are also measured, is violated by construction. This is not an implementation failure but a theoretical limitation: without interventional data or instrumental variables, no observational method can distinguish direct causal effects from confounded associations \citep{pearl2009causality}. The observed compression of hardest-level AUPRC values is therefore expected: confounding removes the clean separation between structural and correlational methods.

\paragraph{Why NOTEARS Dominates Under Density and Feedback.}
NOTEARS combines a linear structural model with an $\ell_1$ penalty and a global acyclicity constraint. This gives it a strong inductive bias when the target graph is sparse or moderately dense: at $\rho=0.3$, it retains AUPRC $0.953$ while the next best method, GES, falls to $0.724$. Under feedback, the acyclicity constraint $h(\bW) = 0$ forces NOTEARS to approximate the best acyclic subgraph of the true (cyclic) structure, which is often close to the original DAG when back-edge magnitudes are small.

\subsection{Pathology Interactions}
\label{sec:interactions}

\begin{figure}[t]
    \centering
    \includegraphics[width=0.90\textwidth]{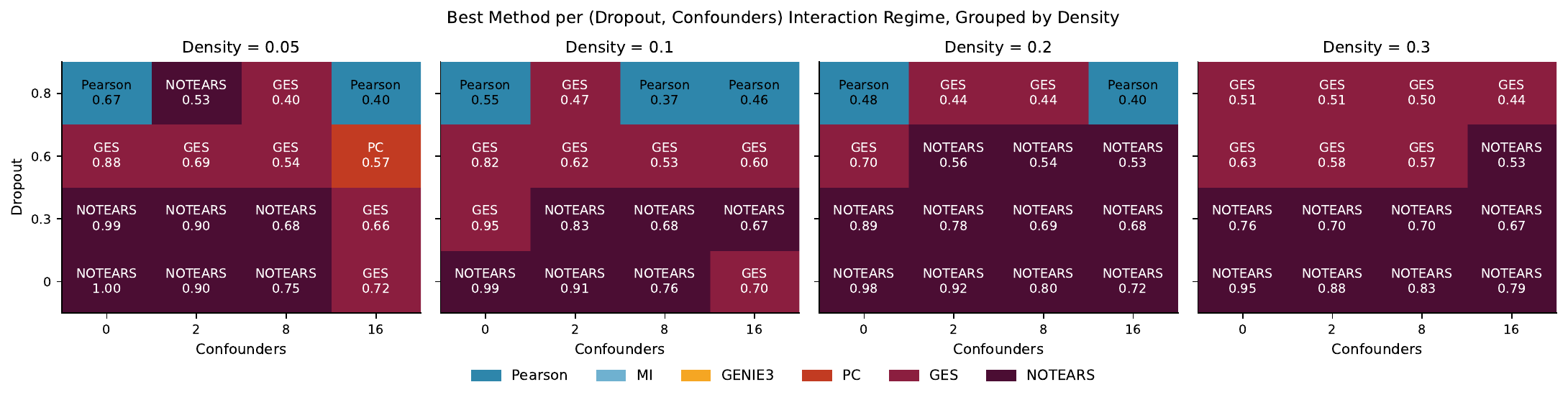}
    \caption{Best method at each $(\delta, k)$-cell, faceted by density $\rho$. Cells are colored by winning method and annotated with mean AUPRC over seeds. NOTEARS owns the sparse low-dropout corner; Pearson takes over once dropout is heavy; GES surfaces along moderate-confounder, high-dropout at higher density.}
    \label{fig:winner_map}
\end{figure}

The single-dial sweeps isolate failure mechanisms but say nothing about how methods behave when multiple pathologies act jointly, the norm rather than the exception in real data. To probe this, we run an interaction sweep over the three most informative dials: dropout $\delta \in \{0, 0.3, 0.6, 0.8\}$, latent confounders $k \in \{0, 2, 8, 16\}$, and network density $\rho \in \{0.05, 0.1, 0.2, 0.3\}$, yielding $64$ joint regimes per method. Figure \ref{fig:winner_map} shows the resulting regime map: at every grid cell, the best-performing method and its mean AUPRC are reported, faceted by density $\rho$ and colored by method class. The full surfaces and per-cell numerical values are reported in Appendix \ref{app:interactions}.

Two findings stand out. First, the regime map is structured but not a clean extrapolation of the single-dial ordering: NOTEARS owns the low-dropout half of the grid across all densities, while the heavy-dropout regime is split between Pearson and GES in a density-conditional way. At low density Pearson wins as the single-dial story predicts, but at $\rho = 0.3$ GES displaces Pearson across the entire $\delta \geq 0.6$ row, a cross-over invisible to single-dial analysis. Second, joint pathology effects are markedly \emph{sub-additive}: the AUPRC drop from the clean corner ($\delta = 0, k = 0, \rho = 0.05$) to the diagonally opposite hardest cell is smaller than the sum of the corresponding single-axis drops for every method. As depicted in Table \ref{tab:interaction} in the Appendix, the gap is largest for MI ($-0.96$) and GENIE3 ($-0.86$), already saturated by dropout alone, and smallest for NOTEARS ($-0.29$).

\section{Discussion}
\label{sec:discussion}

\paragraph{From Rankings to Failure Modes.}
Our controlled experiments provide a mechanistic resolution of benchmark rankings. Different method classes fail for different reasons: causal methods are superior in clean and structurally favorable regimes, dropout selectively damages discretization- and tree-based scores, and confounding compresses differences across all classes. These co-occurring pathologies can make benchmark leaderboards look contradictory, including the familiar cases where simple correlational baselines appear to beat causal methods. Our factorial interaction sweep further shows that the resulting joint failures are largely a sub-additive composition of single-dial mechanisms, so the single-pathology ordering remains a useful reference point.

\paragraph{Practical Guidance.}
Our results yield concrete recommendations: \textit{(i)} under heavy dropout, prefer Pearson correlation or impute before applying causal methods; \textit{(ii)} under suspected confounding, use interventional data; \textit{(iii)} for sparse networks with adequate sample size, NOTEARS provides the best accuracy-per-compute tradeoff (Figure \ref{fig:pareto} in Appendix \ref{app:details}); \textit{(iv)} always examine error decomposition, not just AUPRC, since qualitatively different errors have different downstream consequences.

\paragraph{Limitations.}
Our study has several limitations that should guide interpretation and future work. \emph{(Linear Gaussian SCM)} Our primary analysis uses the same linear additive-noise model assumed by NOTEARS and GES. However, the pathologies we study are not artifacts specific to linearity, and Appendix \ref{app:nonlinear} shows that a nonlinear SCM produces qualitatively similar degradation patterns. \emph{(Scale)} We use $p = 25$ genes, representative of a transcription-factor module but far from genome scale, where complexity may change method rankings. \emph{(Simulation-first)} Our evaluation is entirely synthetic. A systematic study linking simulation pathologies to measurable properties of real perturbational datasets would strengthen the bridge between controlled and real-data evaluation.

\section{Conclusion} \label{sec:conclusion}

We have presented a controlled diagnostic framework for understanding when and why GRN inference methods fail. By isolating seven biologically motivated pathologies and measuring degradation curves for six representative methods, we provide the first systematic attribution of method failure to specific data properties. Our error-type decomposition reveals that aggregate accuracy metrics mask qualitatively different failure modes with distinct biological implications: structural methods fail by hallucinating high-confidence spurious edges, while correlational and tree-based methods are more prone to collapse, reversal, or under-prediction in other regimes. We find that benchmark rankings are best understood through pathology-specific failure modes rather than overall method labels: the widespread observation that \textit{``correlation beats causal''} on real data is one consequence of this broader pattern, driven in particular by dropout and latent confounders that selectively neutralize structural advantages. An interaction sweep over the three most informative pathologies further shows that joint effects compose sub-additively and that cross-over effects can arise (e.g., GES displacing Pearson at high density and heavy dropout) that single-dial analysis alone cannot anticipate. Our framework provides a roadmap for future method development and evaluation, allowing researchers to diagnose new methods against specific pathologies, identify pathologies most in need of methodological innovation (latent confounders remain essentially unsolved from observational data alone), and make informed method choices based on their research goals.

\begin{ack}
This work was supported by the Basque Government under grant DEUSTEK5 -- Human-Centric Computing for Smart Sustainable Communities and Environments (IT1582-22), and by the European Union's Horizon Europe research and innovation programme under the PROTECT-CHILD (Grant Agreement No. 101137423) and LATE-AYA (Grant Agreement No. 101214326) projects.
\end{ack}

\bibliographystyle{plainnat}
\bibliography{References}

@inproceedings{zheng2018dags,
  title={{DAGs with NO TEARS}: Continuous Optimization for Structure Learning},
  author={Zheng, Xun and Aragam, Bryon and Ravikumar, Pradeep K and Xing, Eric P},
  booktitle={Advances in Neural Information Processing Systems},
  volume={31},
  year={2018}
}

@inproceedings{bello2022dagma,
  title={{DAGMA}: Learning {DAGs} via {M}-matrices and a Log-Determinant Acyclicity Characterization},
  author={Bello, Kevin and Aragam, Bryon and Ravikumar, Pradeep},
  booktitle={Advances in Neural Information Processing Systems},
  volume={35},
  year={2022}
}

@book{spirtes2000causation,
  title={Causation, Prediction, and Search},
  author={Spirtes, Peter and Glymour, Clark N and Scheines, Richard},
  year={2000},
  publisher={MIT press}
}

@article{chickering2002optimal,
  title={Optimal Structure Identification with Greedy Search},
  author={Chickering, David Maxwell},
  journal={Journal of Machine Learning Research},
  volume={3},
  pages={507--554},
  year={2002}
}

@book{pearl2009causality,
  title={Causality: Models, Reasoning, and Inference},
  author={Pearl, Judea},
  edition={2nd},
  year={2009},
  publisher={Cambridge University Press}
}

@inproceedings{yu2019dag,
  title={{DAG-GNN}: {DAG} Structure Learning with Graph Neural Networks},
  author={Yu, Yue and Chen, Jie and Gao, Tian and Yu, Mo},
  booktitle={International Conference on Machine Learning},
  pages={7154--7163},
  year={2019},
  organization={PMLR}
}

@inproceedings{lachapelle2020gradient,
  title={Gradient-Based Neural {DAG} Learning},
  author={Lachapelle, S{\'e}bastien and Brouillard, Philippe and Deleu, Tristan and Lacoste-Julien, Simon},
  booktitle={International Conference on Learning Representations},
  year={2020}
}

@inproceedings{brouillard2020differentiable,
  title={Differentiable Causal Discovery from Interventional Data},
  author={Brouillard, Philippe and Lachapelle, S{\'e}bastien and Lacoste, Alexandre and Lacoste-Julien, Simon and Drouin, Alexandre},
  booktitle={Advances in Neural Information Processing Systems},
  volume={33},
  pages={21865--21877},
  year={2020}
}

@article{huynh2010inferring,
  title={Inferring Regulatory Networks from Expression Data Using Tree-Based Methods},
  author={Huynh-Thu, V{\^a}n Anh and Irrthum, Alexandre and Wehenkel, Louis and Geurts, Pierre},
  journal={PLoS ONE},
  volume={5},
  number={9},
  pages={e12776},
  year={2010},
  publisher={Public Library of Science}
}

@article{moerman2019grnboost2,
  title={{GRNBoost2} and {Arboreto}: Efficient and Scalable Inference of Gene Regulatory Networks},
  author={Moerman, Thomas and Aibar Santos, Sara and Bravo Gonz{\'a}lez-Blas, Carmen and Simm, Jaak and Moreau, Yves and Aerts, Jan and Aerts, Stein},
  journal={Bioinformatics},
  volume={35},
  number={12},
  pages={2159--2161},
  year={2019},
  publisher={Oxford University Press}
}

@article{aibar2017scenic,
  title={{SCENIC}: Single-Cell Regulatory Network Inference and Clustering},
  author={Aibar, Sara and Bravo Gonz{\'a}lez-Blas, Carmen and Moerman, Thomas and Huynh-Thu, V{\^a}n Anh and Imrichova, Hana and Hulselmans, Gert and Rambow, Florian and Marine, Jean-Christophe and Geurts, Pierre and Aerts, Jan and others},
  journal={Nature Methods},
  volume={14},
  number={11},
  pages={1083--1086},
  year={2017},
  publisher={Nature Publishing Group}
}

@article{bravo2023scenicPlus,
  title={{SCENIC+}: Single-Cell Multiomic Inference of Enhancers and Gene Regulatory Networks},
  author={Bravo Gonz{\'a}lez-Blas, Carmen and De Winter, Seppe and Hulselmans, Gert and Hecker, Nikolai and Matetovici, Irina and Christiaens, Valerie and Poovathingal, Suresh and Wouters, Jasper and Aibar, Sara and Aerts, Stein},
  journal={Nature Methods},
  volume={20},
  number={9},
  pages={1355--1367},
  year={2023},
  publisher={Nature Publishing Group}
}

@article{shu2021modeling,
  title={Modeling Gene Regulatory Networks Using Neural Network Architectures},
  author={Shu, Hantao and Zhou, Jingtian and Lian, Qiuyu and Li, Han and Zhao, Dan and Zeng, Jianyang and Ma, Jianzhu},
  journal={Nature Computational Science},
  volume={1},
  number={7},
  pages={491--501},
  year={2021},
  publisher={Nature Publishing Group}
}

@article{cui2024scgpt,
  title={{scGPT}: Toward Building a Foundation Model for Single-Cell Multi-omics Using Generative {AI}},
  author={Cui, Haotian and Wang, Chloe and Maan, Hassaan and Pang, Kuan and Luo, Fengning and Duan, Nan and Wang, Bo},
  journal={Nature Methods},
  volume={21},
  number={8},
  pages={1470--1480},
  year={2024},
  publisher={Nature Publishing Group}
}

@article{theodoris2023transfer,
  title={Transfer Learning Enables Predictions in Network Biology},
  author={Theodoris, Christina V and Xiao, Ling and Chopra, Anant and Chaffin, Mark D and Al Sayed, Zeina R and Hill, Matthew C and Mantineo, Helene and Brydon, Elizabeth M and Zeng, Zexian and Liu, X Shirley and others},
  journal={Nature},
  volume={618},
  number={7965},
  pages={616--624},
  year={2023},
  publisher={Nature Publishing Group}
}

@article{dibaeinia2020sergio,
  title={{SERGIO}: A Single-Cell Expression Simulator Guided by Gene Regulatory Networks},
  author={Dibaeinia, Payam and Sinha, Saurabh},
  journal={Cell Systems},
  volume={11},
  number={3},
  pages={252--271},
  year={2020},
  publisher={Elsevier}
}

@article{pratapa2020benchmarking,
  title={Benchmarking Algorithms for Gene Regulatory Network Inference from Single-Cell Transcriptomic Data},
  author={Pratapa, Aditya and Jalihal, Amogh P and Law, Jeffrey N and Bharadwaj, Arun and Murali, T M},
  journal={Nature Methods},
  volume={17},
  number={2},
  pages={147--154},
  year={2020},
  publisher={Nature Publishing Group}
}

@article{prill2010towards,
  title={Towards a Rigorous Assessment of Systems Biology Models: The DREAM3 Challenges},
  author={Prill, Robert J and Marbach, Daniel and Saez-Rodriguez, Julio and Sorger, Peter K and Alexopoulos, Leonidas G and Xue, Xiaowei and Clarke, Neil D and Altan-Bonnet, Gregoire and Stolovitzky, Gustavo},
  journal={PLoS ONE},
  volume={5},
  number={2},
  pages={e9202},
  year={2010},
  publisher={Public Library of Science San Francisco, USA}
}

@article{marbach2012wisdom,
  title={Wisdom of Crowds for Robust Gene Network Inference},
  author={Marbach, Daniel and Costello, James C and K{\"u}ffner, Robert and Vega, Nicole M and Prill, Robert J and Camacho, Diogo M and Allison, Kyle R and {The DREAM5 Consortium} and Kellis, Manolis and Collins, James J and Stolovitzky, Gustavo},
  journal={Nature Methods},
  volume={9},
  number={8},
  pages={796--804},
  year={2012},
  publisher={Nature Publishing Group}
}

@article{chevalley2025causalbench,
  title={{CausalBench}: A Large-Scale Benchmark for Network Inference from Single-Cell Perturbation Data},
  author={Chevalley, Mathieu and Roohani, Yusuf H and Mehrjou, Arash and Leskovec, Jure and Schwab, Patrick},
  journal={Communications Biology},
  volume={8},
  number={1},
  pages={412},
  year={2025},
  publisher={Nature Publishing Group UK London}
}

@article{nourisa2025genernib,
  title={{geneRNIB}: A Living Benchmark for Gene Regulatory Network Inference},
  author={Nourisa, Jalil and Passemiers, Antoine and Stock, Marco and Zeller-Plumhoff, Berit and Cannoodt, Robrecht and Arnold, Christian and Tong, Alexander and Hartford, Jason and Scialdone, Antonio and Moreau, Yves and Li, Yang and Luecken, Malte D},
  journal={bioRxiv},
  year={2025},
  doi={10.1101/2025.02.25.640181}
}

@article{replogle2022mapping,
  title={Mapping Information-Rich Genotype--Phenotype Landscapes with Genome-Scale {Perturb-seq}},
  author={Replogle, Joseph M and Saunders, Reuben A and Pogson, Angela N and Hussmann, Jeffrey A and Lenail, Alexander and Guna, Alina and Mascibroda, Lauren and Wagner, Eric J and Adelman, Karen and Lithwick-Yanai, Gila and others},
  journal={Cell},
  volume={185},
  number={14},
  pages={2559--2575},
  year={2022},
  publisher={Elsevier}
}

@article{roohani2024predicting,
  title={Predicting Transcriptional Outcomes of Novel Multigene Perturbations with {GEARS}},
  author={Roohani, Yusuf and Huang, Kexin and Leskovec, Jure},
  journal={Nature Biotechnology},
  volume={42},
  number={6},
  pages={927--935},
  year={2024},
  publisher={Nature Publishing Group}
}

@article{ahlmann2025deep,
  title={Deep-Learning-Based Gene Perturbation Effect Prediction Does Not Yet Outperform Simple Linear Baselines},
  author={Ahlmann-Eltze, Constantin and Huber, Wolfgang and Anders, Simon},
  journal={Nature Methods},
  volume={22},
  number={8},
  pages={1657--1661},
  year={2025},
  publisher={Nature Publishing Group}
}

@inproceedings{lopez2022large,
  title={Large-Scale Differentiable Causal Discovery of Factor Graphs},
  author={Lopez, Romain and H{\"u}tter, Jan-Christian and Pritchard, Jonathan K and Regev, Aviv},
  booktitle={Advances in Neural Information Processing Systems},
  volume={35},
  pages={19290--19303},
  year={2022}
}

@book{cover1991elements,
  title={Elements of Information Theory},
  author={Cover, Thomas M and Thomas, Joy A},
  year={1991},
  publisher={Wiley-Interscience}
}

@article{kamimoto2023dissecting,
  title={Dissecting Cell Identity via Network Inference and In Silico Gene Perturbation},
  author={Kamimoto, Kenji and Stringa, Blerta and Hoffmann, Christy M and Jindal, Kunal and Solnica-Krezel, Lilianna and Morris, Samantha A},
  journal={Nature},
  volume={614},
  number={7949},
  pages={742--751},
  year={2023},
  publisher={Nature Publishing Group}
}

@inproceedings{nazaret2024stable,
  title={Stable Differentiable Causal Discovery},
  author={Nazaret, Achille and Hong, Justin and Azizi, Elham and Blei, David},
  booktitle={Proceedings of the 41st International Conference on Machine Learning},
  pages={37413--37445},
  year={2024}
}

@article{kalfon2025scprint,
  title={{scPRINT}: Pre-Training on 50 Million Cells Allows Robust Gene Network Predictions},
  author={Kalfon, J{\'e}r{\'e}mie and Samaran, Jules and Peyr{\'e}, Gabriel and Cantini, Laura},
  journal={Nature Communications},
  volume={16},
  number={1},
  pages={3607},
  year={2025},
  publisher={Nature Publishing Group}
}

\newpage
\appendix

\section{Inference Method Details}
\label{app:methods}

We provide complete descriptions of the six inference methods evaluated in this study, including their algorithmic details, hyperparameter choices, and key assumptions.

\paragraph{Pearson Correlation.}
Pearson correlation computes the normalized covariance between every pair of genes: $S_{ij} = |\mathrm{corr}(X_{\cdot i}, X_{\cdot j})|$. The absolute value is taken so that both positive and negative associations contribute equally to the score. The result is a symmetric score matrix with entries in $[0, 1]$. This method makes no structural assumptions and is relatively robust in our dropout sweep because zero-inflation attenuates many pairwise magnitudes without destroying the entire score ranking. Its main limitation is the inability to distinguish direct edges from indirect associations or to recover edge orientation.

\paragraph{Mutual Information (MI).}
Mutual information is estimated by discretizing each gene's expression into equal-frequency bins and computing the empirical joint entropy: $S_{ij} = H(X_{\cdot i}) + H(X_{\cdot j}) - H(X_{\cdot i}, X_{\cdot j})$. We use 6 bins per gene, as in standard GRN benchmarks \citep{pratapa2020benchmarking}. The resulting score matrix is symmetric. MI can in principle detect nonlinear dependencies that Pearson misses, but the discretization step is highly sensitive to zero-inflation: at high dropout, large blocks of tied zero values make the empirical ranks and joint entropies unstable, explaining MI's catastrophic collapse under the dropout pathology.

\paragraph{GENIE3.}
GENIE3 \citep{huynh2010inferring} casts GRN inference as a feature-importance problem. For each target gene $j$, a Random Forest regressor is trained on all other genes $\{X_{\cdot i}\}_{i \neq j}$ to predict $X_{\cdot j}$, and the feature importances are used as asymmetric edge scores: $S_{ij} = \mathrm{importance}(X_{\cdot i} \to X_{\cdot j})$. We use 50 trees per forest with default Scikit-Learn hyperparameters. The resulting score matrix is asymmetric, with $S_{ij}$ encoding how predictive gene $i$ is for gene $j$. GENIE3 is robust to nonlinear relationships but, like MI, relies on tree-based splits that become unreliable under heavy dropout. Under dense networks, the predictive signal is spread across many regulators, reducing individual importances.

\paragraph{PC (Peter--Clark).}
The PC algorithm \citep{spirtes2000causation} performs constraint-based causal discovery by iteratively removing edges from the complete graph based on conditional independence tests. We use Fisher $z$-tests at significance level $\alpha = 0.05$ with a maximum conditioning-set size of 2. The output is a completed partially directed acyclic graph (CPDAG), from which we extract the skeleton (undirected edges) weighted by marginal Pearson correlation magnitude as the score matrix. This makes the effective score matrix symmetric. PC assumes causal sufficiency (no latent confounders) and faithfulness; violation of either leads to spurious or missing edges. With a conditioning-set limit of 2, PC is computationally efficient but may fail to remove edges in high-density settings where longer adjustment paths are needed.

\paragraph{GES (Greedy Equivalence Search).}
GES \citep{chickering2002optimal} is represented by a compact GES-like forward search using a local linear-Gaussian BIC score. We greedily add parents that improve the target's BIC, limit the maximum number of parents per node to 3, and restrict candidates using a variance-based proxy topological ordering to enforce acyclicity. The resulting asymmetric score matrix records each selected parent's positive delta-BIC contribution. This implementation captures the score-based, sparsity-constrained behavior we want to test, but it is not a full two-phase CPDAG search. In dense settings, the BIC score may not strongly discriminate among indirect dependencies, causing confounded associations to be included.

\paragraph{NOTEARS.}
NOTEARS \citep{zheng2018dags} reformulates DAG learning as a continuous optimization problem. It minimizes the penalized least-squares objective:
\begin{equation}
    \min_{\bW}\; \frac{1}{2n}\|\bX - \bX\bW\|_F^2 + \lambda\|\bW\|_1 \quad \text{s.t.}\quad h(\bW) = \mathrm{tr}(e^{\bW \circ \bW}) - p = 0
\end{equation}
where $h(\bW) = 0$ is a differentiable algebraic constraint that is zero if and only if $\bW$ is a DAG. We use $\lambda = 0.05$ and threshold learned coefficients at $|W_{ij}| < 0.1$ to obtain the final edge set. The resulting weight matrix is asymmetric and encodes both direction and magnitude. The $\ell_1$ penalty promotes sparsity and suppresses indirect associations, providing the strongest inductive bias for direct causal edge recovery in sparse, low-noise networks. The acyclicity constraint causes NOTEARS to prefer the best acyclic approximation when feedback loops are present, which often recovers the dominant causal direction.

\section{Additional Linear SCM Results}
\label{app:linear_extra}

We provide additional analyses of the linear SCM results, including method family comparisons, directed AUPRC degradation curves, and baseline AUPRC values under clean conditions.
\paragraph{Method Family Comparison}
\label{app:families}

\begin{figure}[t]
    \centering
    \includegraphics[width=\textwidth]{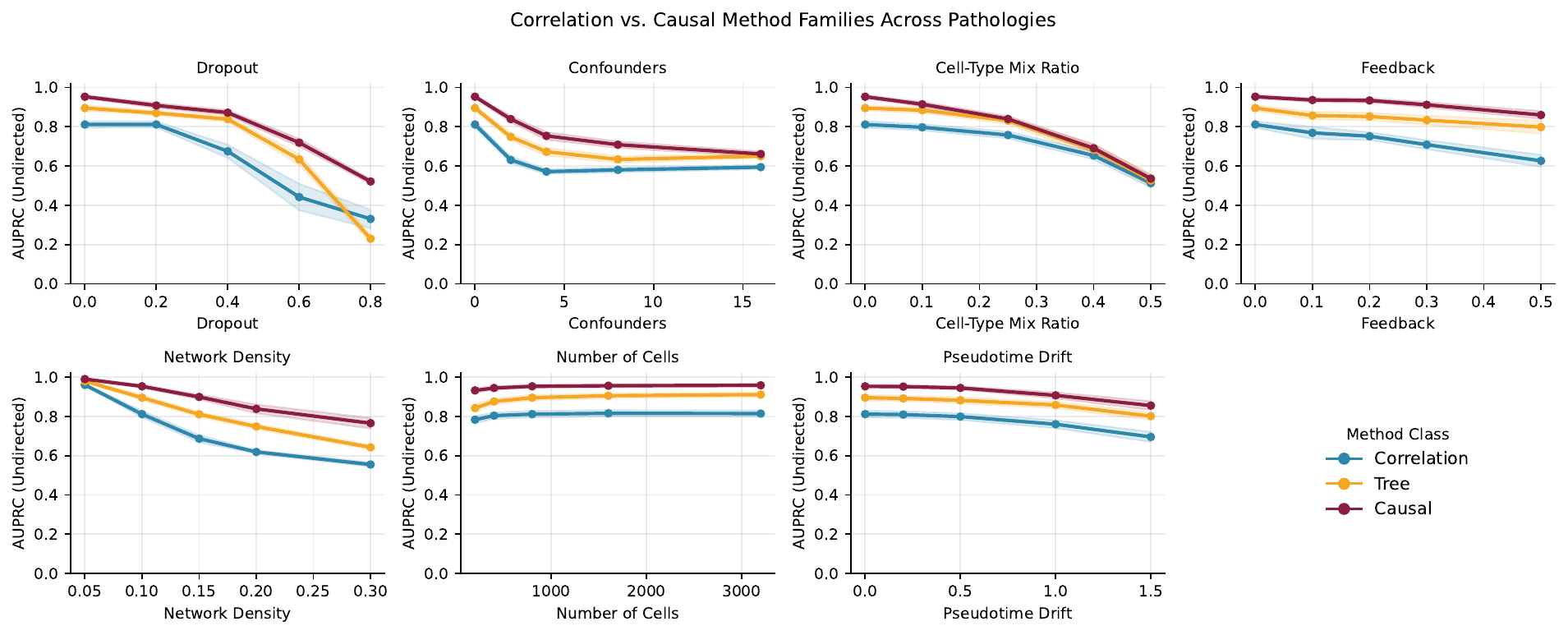}
    \caption{Method family comparison: mean $\pm$ SEM across methods within each family. The causal family has the highest average AUPRC across all sweeps, with the clearest advantage under density, feedback, and sample-size variation.}
    \label{fig:families}
\end{figure}

Figure \ref{fig:families} aggregates individual methods into families: correlation (Pearson, MI), tree-ensemble (GENIE3), and causal (PC, GES, NOTEARS). The causal family has the highest mean undirected AUPRC across all seven sweeps, but the margin is pathology-dependent: it is largest under density, feedback, and sample-size variation, and smallest when confounding or cell-type mixing compresses all methods. Notably, the causal family's advantage under the density sweep is large: at $\rho = 0.3$, the causal family mean is $0.77$ AUPRC while the correlation family drops to $\approx 0.55$. Conversely, at peak dropout ($\delta = 0.8$), MI's collapse pulls the correlation mean below $0.4$, obscuring the fact that Pearson alone remains competitive. Disaggregating by method (Figure \ref{fig:headline}) is therefore essential for interpreting family-level aggregates.

\paragraph{Directed AUPRC Degradation Curves}
\label{app:directed}

\begin{figure}[t]
    \centering
    \includegraphics[width=\textwidth]{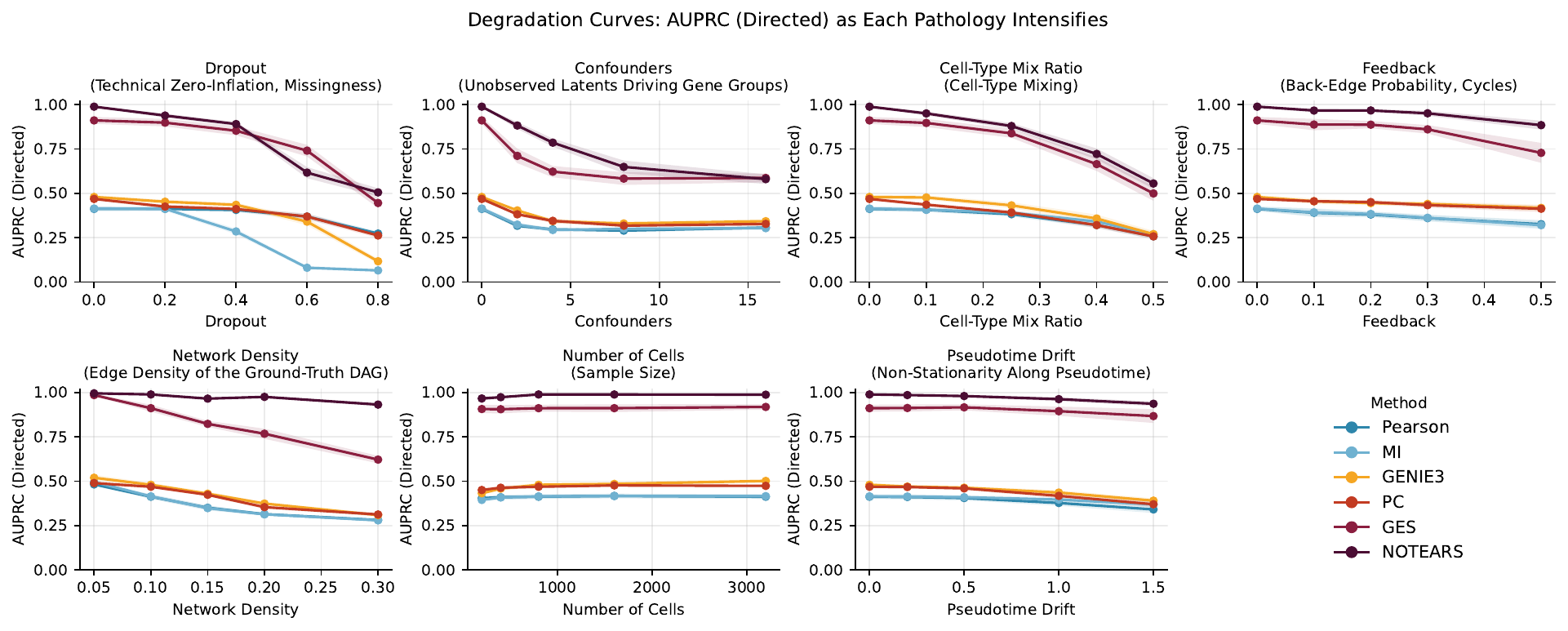}
    \caption{Directed AUPRC degradation across all seven pathologies. Symmetric score matrices (Pearson, MI, and the PC skeleton implementation) lose orientation information and operate near their random baseline in directed evaluation, whereas GES and NOTEARS maintain strong directed performance under clean and moderately perturbed conditions.}
    \label{fig:headline_directed}
\end{figure}

\begin{figure}[t]
    \centering
    \includegraphics[width=\textwidth]{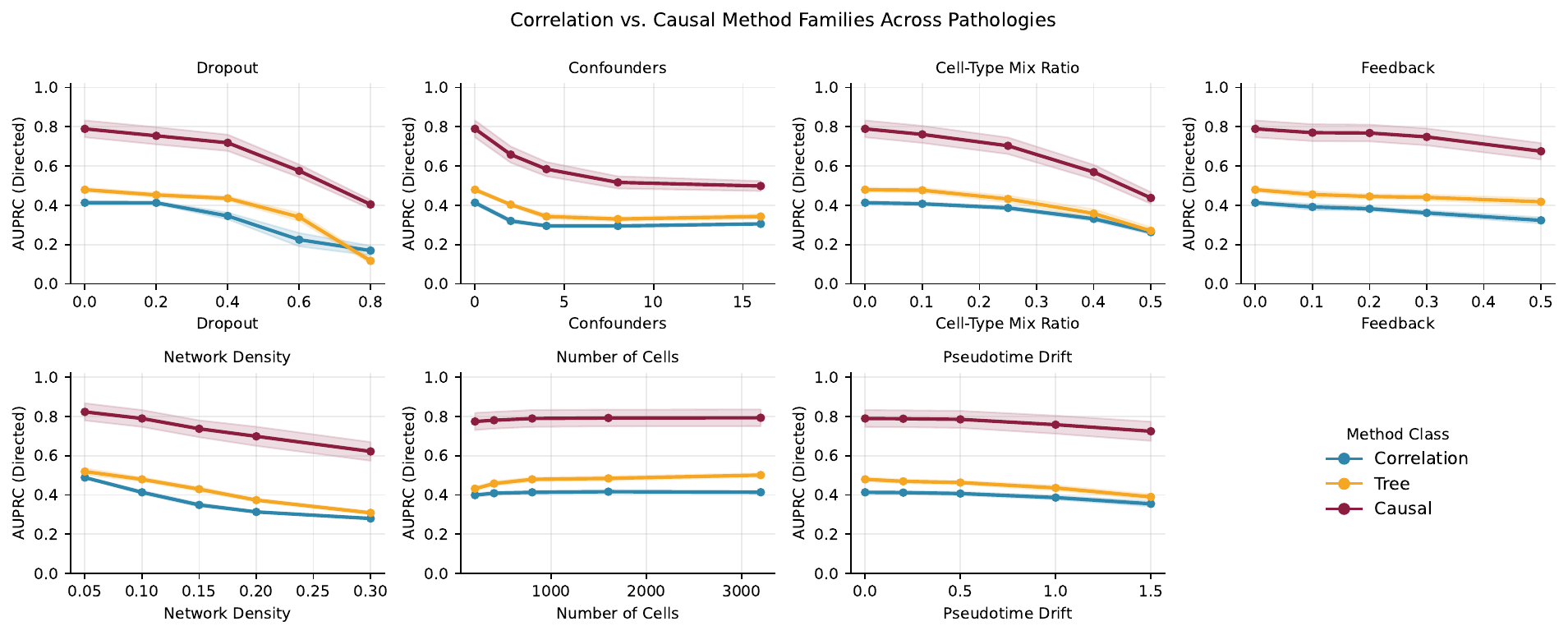}
    \caption{Directed AUPRC method family comparison. The causal family advantage is substantially larger in directed evaluation because GES and NOTEARS produce asymmetric parent-to-target scores, while correlation and tree-ensemble methods either produce symmetric scores or inherit directional limitations from their feature-importance formulation.}
    \label{fig:families_directed}
\end{figure}

All main-text analyses use undirected AUPRC to compare edge recovery without penalizing symmetric score matrices. Figure \ref{fig:headline_directed} repeats the degradation analysis with directed AUPRC. The main difference is expected: Pearson, MI, and PC lose roughly half their clean AUPRC because their scores are symmetric (producing tied scores in both directions), while GES and NOTEARS retain directed AUPRC close to their undirected values under clean conditions ($0.911$ and $0.989$, respectively; see Table \ref{tab:baseline}). Importantly, the \emph{relative} degradation patterns across pathologies are preserved: dropout still collapses MI and GENIE3, confounding compresses method differences, and NOTEARS dominates under the density and feedback pathologies. The absolute gap between structural and correlational methods is wider in the directed setting, as edge orientation is precisely what symmetric methods cannot recover. Figure \ref{fig:families_directed} shows the corresponding family-level directed comparison, where the causal family retains a larger margin over the correlation family due to GES and NOTEARS's asymmetric parent-to-target scoring.

\paragraph{Baseline AUPRC Values}
\label{app:baseline}

Table \ref{tab:baseline} reports the AUPRC values under clean conditions (default pathology levels) for reference. These baseline values establish the upper bound on performance for each method and represent the regime most favorable to causal discovery: a linear Gaussian SCM with adequate sample size, no dropout, no latent confounders, and moderate network density.

\begin{table}[t]
\centering
\caption{Baseline AUPRC under default settings ($p = 25$ genes, $n = 800$ cells, $\rho = 0.1$, no pathologies), linear Gaussian SCM. Mean over 10 seeds.}
\label{tab:baseline}
\vspace{0.5em}
\small
\begin{tabular}{@{}lcccccc@{}}
\toprule
& \textbf{Pearson} & \textbf{MI} & \textbf{GENIE3} & \textbf{PC} & \textbf{GES} & \textbf{NOTEARS} \\
\midrule
AUPRC (undirected) & 0.811 & 0.813 & 0.895 & 0.923 & 0.944 & 0.992 \\
AUPRC (directed)   & 0.412 & 0.414 & 0.479 & 0.468 & 0.911 & 0.989 \\
\bottomrule
\end{tabular}
\end{table}

The large gap between undirected and directed AUPRC for Pearson, MI, and PC reflects their symmetric score matrices: without orientation signal, these methods are limited to approximately chance-level directed AUPRC relative to the number of ordered pairs. GENIE3's moderate directed AUPRC ($0.479$) reflects partial asymmetry from its feature-importance formulation, but this is far below its undirected performance because importance scores do not reliably encode causal direction. GES and NOTEARS maintain near-perfect directed AUPRC ($> 0.9$), demonstrating that their asymmetric parent-to-target scores recover edge orientation reliably under clean conditions. This clean-condition baseline is the reference against which all degradation in the pathology sweeps should be interpreted.

\section{Nonlinear SCM Results}
\label{app:nonlinear}

To test whether the diagnostic conclusions depend on the linear Gaussian SCM, we replicate the full sweep under a nonlinear SCM where each gene's expression is a nonlinear function of its parents plus noise. We choose a simple nonlinearity, the hyperbolic tangent (\textit{tanh}), which preserves the additive noise structure while introducing bounded nonlinear interactions that are common in gene regulation.

\begin{definition}[Nonlinear SCM (Tanh)]
\label{def:nonlinear}
The nonlinear variant replaces the linear structural assignment with:
\begin{equation}
    X_j^{(c)} = \tanh\!\left(\sum_{i \in \mathcal{P}_\mathcal{G}(j)} W_{ij}\, X_i^{(c)}\right) + \varepsilon_j^{(c)}, \quad \varepsilon_j^{(c)} \overset{\text{iid}}{\sim} \mathcal{N}(0, \sigma^2)
\end{equation}
For acyclic graphs, sampling proceeds by ancestral ordering. For cyclic graphs (with feedback), we use fixed-point iteration $\bX^{(t+1)} = \tanh(\bX^{(t)} \bW) + \bE$, which converges when $\|\bW\|_2 < 1$ (guaranteed by our rescaling) since the tanh nonlinearity is 1-Lipschitz.
\end{definition}

\begin{figure}[t]
    \centering
    \includegraphics[width=\textwidth]{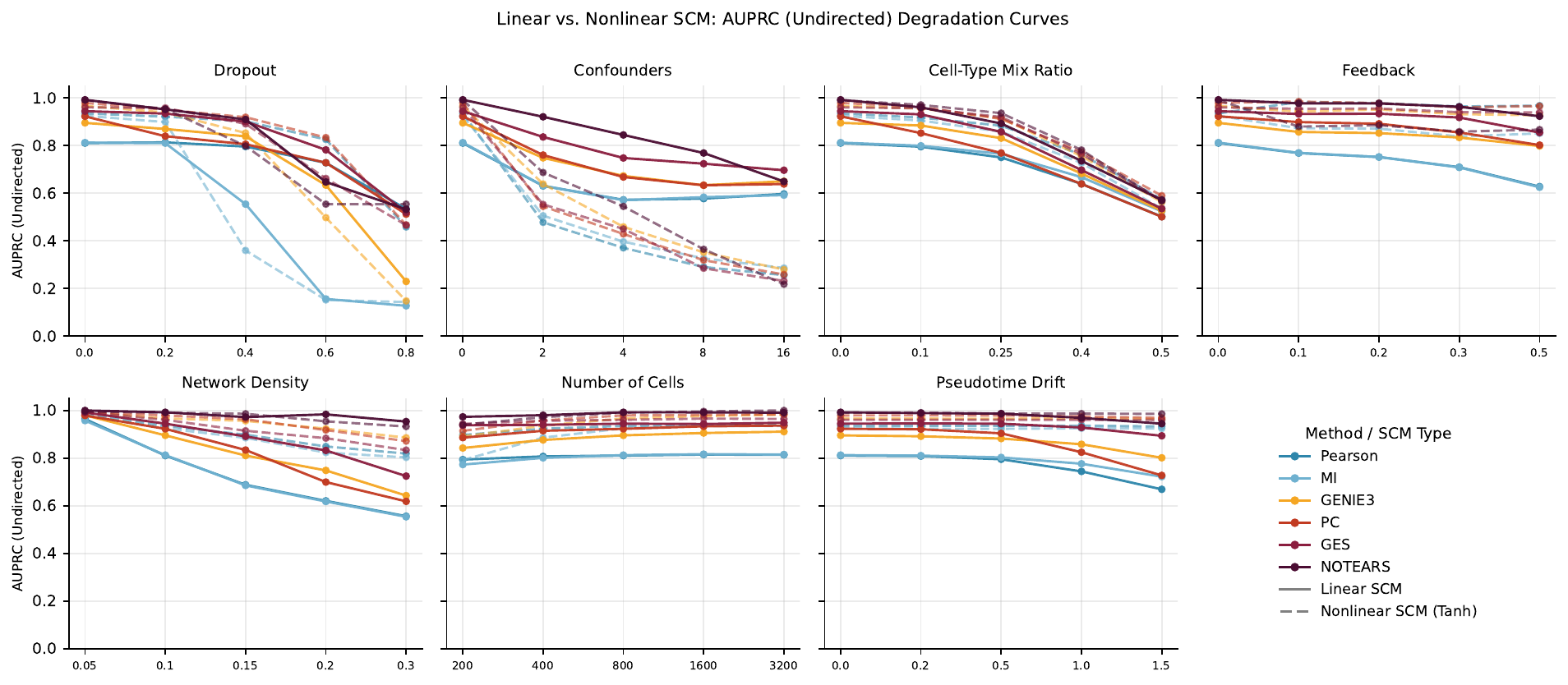}
    \caption{Linear (solid) vs. nonlinear tanh (dashed) SCM, using undirected AUPRC. The nonlinear SCM changes absolute difficulty and some method rankings, but the diagnostic stressors remain visible: dropout still collapses MI and GENIE3, confounding remains broadly damaging, and NOTEARS remains strongest under high density.}
    \label{fig:nonlinear}
\end{figure}

\begin{figure}[t]
    \centering
    \includegraphics[width=\textwidth]{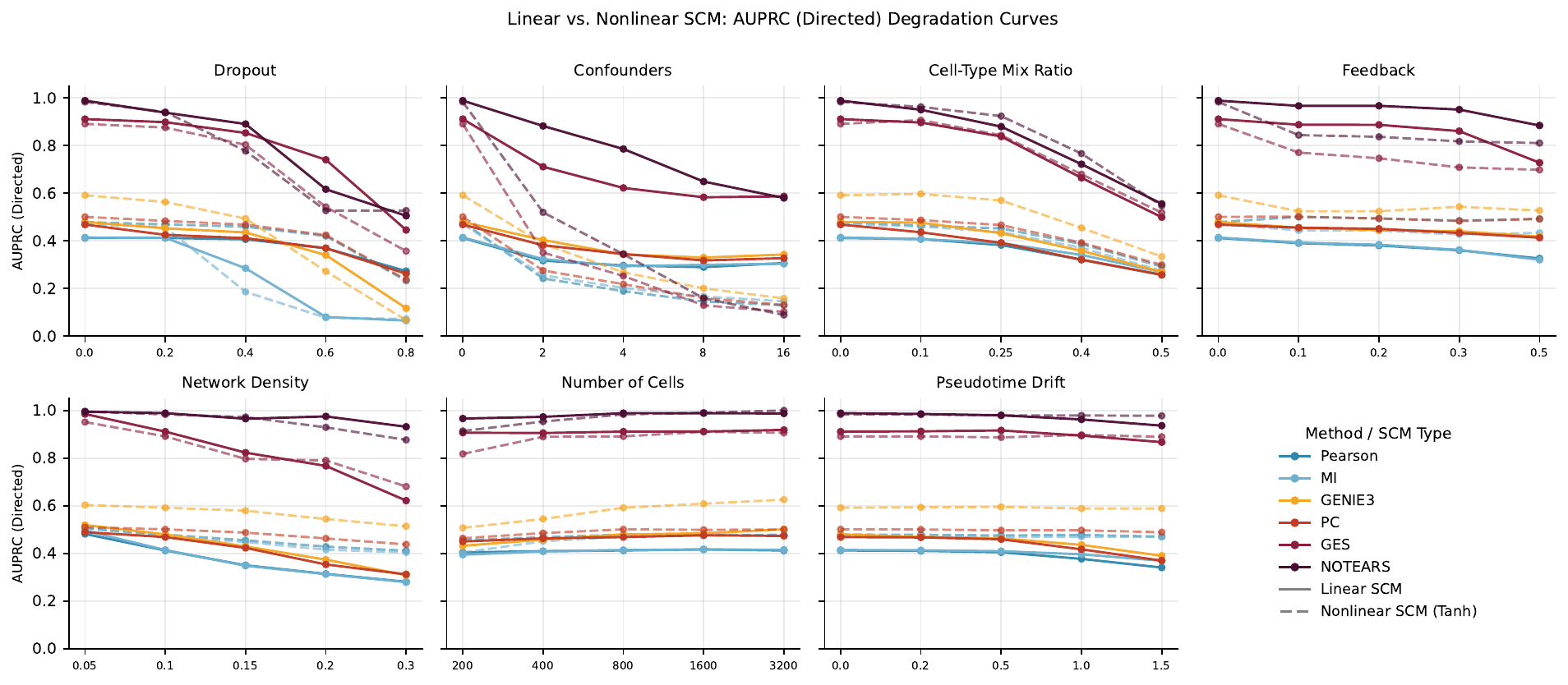}
    \caption{Linear vs. nonlinear SCM comparison using directed AUPRC. GES and NOTEARS retain the strongest orientation signal in clean and moderately perturbed settings, while symmetric methods remain limited by tied scores in both directions. The directed gap between structural and correlational methods is wider than in the undirected case, as orientation recovery is the primary differentiator between method classes.}
    \label{fig:nonlinear_directed}
\end{figure}

Figure \ref{fig:nonlinear} overlays linear (solid) and nonlinear (dashed) degradation curves. The nonlinear sweep should not be read as an identical replay of the linear results. Because the tanh nonlinearity bounds expression values, several settings become easier in absolute undirected AUPRC (e.g., NOTEARS reaches near-perfect AUPRC under clean conditions in both regimes), while strong latent confounding becomes harder for all methods due to the increased expressiveness of nonlinear confounding pathways. Critically, the qualitative ordering of pathology severity is preserved: dropout remains the most destructive pathology for MI and GENIE3 (whose discretization-based scores are equally disrupted under either SCM); confounding compresses method differences regardless of the functional form; and NOTEARS retains the largest advantage in the high-density regime, where its sparsity-inducing $\ell_1$ penalty and acyclicity constraint continue to suppress indirect associations. The nonlinear results thus corroborate the central diagnostic conclusions and demonstrate that our findings are not artifacts of the linear Gaussian modeling assumption. Figure \ref{fig:nonlinear_directed} shows the corresponding directed comparison, where the gap between structural and symmetric methods is further amplified because the nonlinearity does not change the fundamental inability of Pearson, MI, and PC to recover edge orientation.

Table \ref{tab:baseline_nonlinear} reports the corresponding baseline AUPRC under the nonlinear SCM at the same default settings. Compared to the linear baseline (Table \ref{tab:baseline}), the bounded tanh nonlinearity increases clean undirected AUPRC for several methods, while NOTEARS retains near-identical performance. Directed AUPRC patterns are preserved: GES and NOTEARS maintain a large gap over symmetric methods, with GENIE3's directed score improving due to the saturating nonlinearity making parent contributions more distinguishable from indirect associations.

\begin{table}[t]
\centering
\caption{Baseline AUPRC under default settings ($p = 25$ genes, $n = 800$ cells, $\rho = 0.1$, no pathologies), nonlinear tanh SCM. Mean over 10 seeds.}
\label{tab:baseline_nonlinear}
\vspace{0.5em}
\small
\begin{tabular}{@{}lcccccc@{}}
\toprule
& \textbf{Pearson} & \textbf{MI} & \textbf{GENIE3} & \textbf{PC} & \textbf{GES} & \textbf{NOTEARS} \\
\midrule
AUPRC (undirected) & 0.935 & 0.925 & 0.967 & 0.979 & 0.960 & 0.991 \\
AUPRC (directed)   & 0.479 & 0.470 & 0.591 & 0.501 & 0.891 & 0.983 \\
\bottomrule
\end{tabular}
\end{table}

\section{Pathology Interaction Sweeps}
\label{app:interactions}

This appendix expands on the factorial interaction analysis introduced in Section \ref{sec:interactions}. We jointly vary the three most informative single-dial pathologies, dropout $\delta \in \{0, 0.3, 0.6, 0.8\}$, latent confounders $k \in \{0, 2, 8, 16\}$, and network density $\rho \in \{0.05, 0.1, 0.2, 0.3\}$, on a fully crossed grid ($64$ joint regimes), with all remaining pathology dials held at their benign defaults and $5$ random seeds per cell. The fixed simulator settings match the single-dial sweeps so that interaction-specific effects are not confounded with scale.

\begin{figure}[t]
    \centering
    \includegraphics[width=\textwidth]{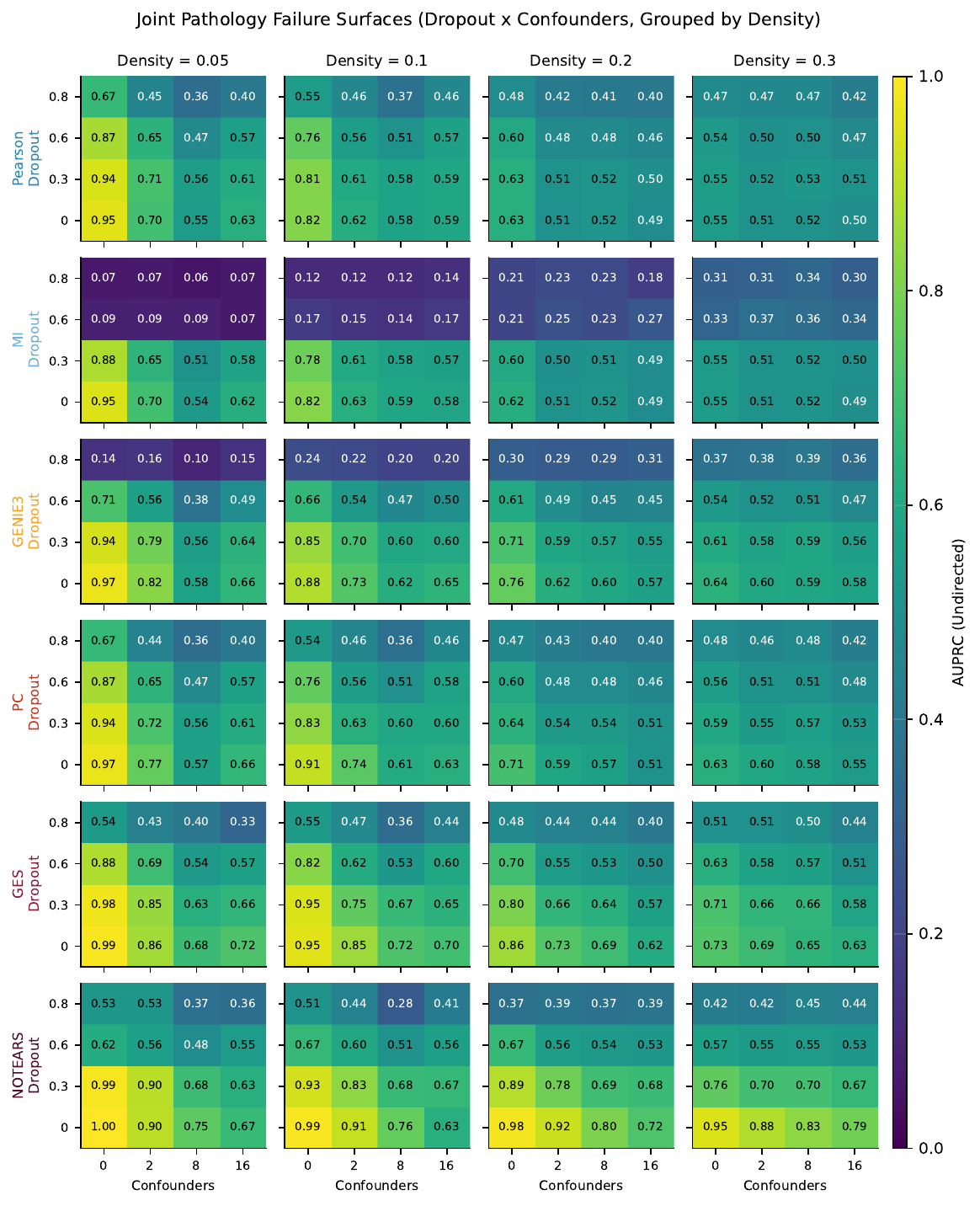}
    \caption{Per-method failure surfaces over the dropout $\times$ confounders plane, faceted by density. Each cell reports mean undirected AUPRC over $5$ seeds. Shared $[0,1]$ colour scale. The dropout axis dominates degradation for MI and GENIE3, while NOTEARS retains strong performance in the low-dropout regime before dropping sharply at higher dropout.}
    \label{fig:interaction_surfaces}
\end{figure}

\paragraph{Per-Method Failure Surfaces.}
Figure \ref{fig:interaction_surfaces} shows, for each method, the mean undirected AUPRC over the dropout $\times$ confounders plane at each density slice. Three patterns are visible. First, the dropout axis is the dominant degrader for MI and GENIE3 across every density; their surfaces collapse uniformly along this axis with little additional movement once $\delta > 0.6$. Second, NOTEARS retains a high-AUPRC region at low dropout across confounder levels, but this region contracts once dropout reaches $\delta = 0.6$ and is largely lost at $\delta = 0.8$. Third, density modulates absolute AUPRC for every method except NOTEARS, whose ridge stays near $1.0$ across the entire density axis; PC and GES also respond to density less than the correlation- and tree-based methods, consistent with Table \ref{tab:delta}.

\paragraph{Sub-Additivity of Joint Pathologies.}
Table \ref{tab:interaction} quantifies the sub-additivity claim from Section \ref{sec:interactions}. For each method we compute the AUPRC drop from the cleanest joint cell ($\delta = 0$, $k = 0$, $\rho = 0.05$) to the worst joint cell ($\delta = 0.8$, $k = 16$, $\rho = 0.3$), and compare it against the sum of the corresponding single-dial drops. The interaction term, defined as $\Delta_{\text{joint}} - \Delta_{\text{additive}}$, is negative for every method, indicating that pathologies share failure modes rather than compounding independently. The effect is largest for MI and GENIE3 (already saturated by dropout alone) and smallest for NOTEARS, whose degradation is closest to additive.

\begin{table}[t]
\centering
\caption{Joint vs.\ additive AUPRC degradation across the dropout $\times$ confounders $\times$ density grid. $\Delta_{\text{joint}}$ is the AUPRC drop from the cleanest to the worst joint cell; $\Delta_d$, $\Delta_k$, $\Delta_\rho$ are the corresponding single-dial drops along each axis (other dials held at their cleanest level); $\Delta_{\text{add}} = \Delta_d + \Delta_k + \Delta_\rho$; interaction $= \Delta_{\text{joint}} - \Delta_{\text{add}}$. All entries are AUPRC differences (lower means more degradation); negative interaction values indicate sub-additive composition.}
\label{tab:interaction}
\vspace{0.5em}
\small
\setlength{\tabcolsep}{4.5pt}
\begin{tabular}{@{}lcccccc@{}}
\toprule
\textbf{Method} & $\mathbf{AUPRC_0}$ & $\mathbf{\Delta_{joint}}$ & $\mathbf{\Delta_d}$ & $\mathbf{\Delta_k}$ & $\mathbf{\Delta_\rho}$ & \textbf{Interaction} \\
\midrule
Pearson  & 0.954 & 0.532 & 0.282 & 0.326 & 0.402 & $-0.477$ \\
MI       & 0.949 & 0.646 & 0.878 & 0.326 & 0.400 & $-0.958$ \\
GENIE3   & 0.971 & 0.613 & 0.829 & 0.313 & 0.328 & $-0.857$ \\
PC       & 0.972 & 0.553 & 0.304 & 0.309 & 0.340 & $-0.399$ \\
GES      & 0.989 & 0.549 & 0.444 & 0.267 & 0.259 & $-0.421$ \\
NOTEARS  & 1.000 & 0.563 & 0.470 & 0.328 & 0.052 & $-0.287$ \\
\bottomrule
\end{tabular}
\end{table}

\paragraph{Reading the Winner Map.}
The regime map in Figure \ref{fig:winner_map} (Section \ref{sec:interactions}) summarizes the surfaces by reporting only the best method at each cell. NOTEARS owns the low-dropout half of the grid across all densities, with the largest margin in the sparser slices. The heavy-dropout regime is more nuanced: at low density Pearson wins, consistent with its graceful degradation under MNAR zero-inflation, but as density increases GES progressively reclaims those cells, and at $\rho = 0.3$ GES displaces Pearson entirely across the $\delta \geq 0.6$ rows. This cross-over is the most visible deviation from the single-dial ordering and is consistent with GES's forward-search BIC selection benefiting from richer dependency structure once edges are abundant.

\section{Experimental Details}
\label{app:details}

\paragraph{Reproducibility.}
All experiments use NumPy's default random number generator with seeds $\{0, 1, \ldots, 9\}$. The full experiment suite can be reproduced with a single command and run on a standard CPU with consumer-grade specifications. Code and data will be made publicly available upon publication.

\paragraph{Graph Construction.}
Ground-truth DAGs are constructed with node ordering $0 < 1 < \cdots < p-1$ as the topological order. Full details are provided in Section \ref{sec:dgp}.

\paragraph{Software.}
Implementation uses NumPy, SciPy, Scikit-Learn (random forests for GENIE3), NetworkX (graph operations), and Matplotlib (visualization). No GPU is required.

\begin{figure}[H]
    \centering
    \includegraphics[width=0.7\textwidth]{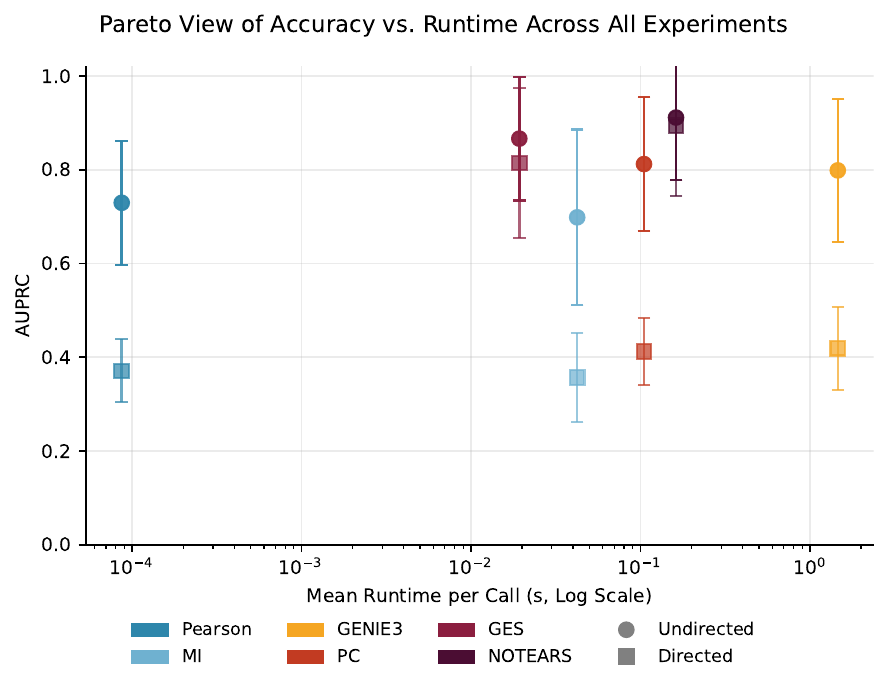}
    \caption{Pareto view of accuracy vs.\ runtime (log scale), averaged across all linear-SCM experiments. \textit{Circles:} undirected AUPRC; \textit{Squares:} directed AUPRC. Pearson is fastest ($\sim 10^{-4}$s) but less accurate; GES is both fast ($\sim 2\times10^{-2}$s) and structurally accurate; NOTEARS gives the highest average AUPRC at moderate cost ($\sim 1.6\times10^{-1}$s); GENIE3 is slowest ($\sim 1.5$s) without a commensurate accuracy gain.}
    \label{fig:pareto}
\end{figure}

\paragraph{Accuracy--Runtime Tradeoff.}
Figure \ref{fig:pareto} presents the accuracy--runtime Pareto frontier, averaged across all linear-SCM experiments. Runtimes span more than four orders of magnitude: Pearson correlation completes in microseconds ($\sim 10^{-4}$s) while GENIE3's random-forest fitting takes $\sim 1.5$s. NOTEARS occupies a compelling middle ground, achieving the highest average undirected AUPRC ($0.912$ averaged across all pathology levels) at a moderate cost of $\sim 0.16$s per experiment. GES is the most efficient structural method, retaining strong directed AUPRC under clean conditions at roughly one-tenth of NOTEARS's runtime. The Pareto analysis reinforces the practical recommendation from the main text: for sparse networks with adequate sample size and without heavy dropout or confounding, NOTEARS provides the best accuracy-per-compute tradeoff. For settings where runtime is the primary constraint, GES is a strong alternative. Pearson correlation, despite being the fastest method by a wide margin, is dominated in accuracy by all structural methods under clean conditions, and should only be preferred when dropout is heavy (Section \ref{sec:diagnoses}).

\clearpage
\newpage

\end{document}